\documentclass[sigconf]{acmart}
\acmISBN{}
\acmDOI{}
\acmConference[ICEA2024]{2024 International Conference on Intelligent Computing and its Emerging Applications}{November 28 – 29, 2024}{Tokyo, Japan}

\usepackage{caption}
\usepackage{subcaption}
\usepackage{bbding}

\AtBeginDocument{%
  }

\setcopyright{acmcopyright}
\copyrightyear{2024}
\acmYear{2024}
\usepackage[ruled,vlined]{algorithm2e}

\begin{document}

\title{Neural Error Covariance Estimation for Precise LiDAR Localization}

\author{Minoo Dolatabadi$^1$, Fardin Ayar$^1$, Ehsan Javanmardi$^2$, Manabu Tsukada$^2$, Mahdi Javanmardi$^1$}
\email{minoo.dolatabadi@aut.ac.ir,fardin.ayar@aut.ac.ir, ejavanmardi@g.ecc.u-tokyo.ac.jp, mtsukada@g.ecc.u-tokyo.ac.jp, mjavan@aut.ac.ir}
\affiliation{%
\institution{$^1$Department of Computer Engineering, Amirkabir University of Technology, Tehran, Iran \\
$^2$Graduate School of Information Science and Technology, The University of Tokyo, Tokyo, Japan}
\country{}}

\begin{abstract}
Autonomous vehicles have gained significant attention due to technological advancements and their potential to transform transportation. A critical challenge in this domain is precise localization, particularly in LiDAR-based map matching, which is prone to errors due to degeneracy in the data. Most sensor fusion techniques, such as the Kalman filter, rely on accurate error covariance estimates for each sensor to improve localization accuracy. However, obtaining reliable covariance values for map matching remains a complex task. To address this challenge, we propose a neural network-based framework for predicting localization error covariance in LiDAR map matching. To achieve this, we introduce a novel dataset generation method specifically designed for error covariance estimation. In our evaluation using a Kalman filter, we achieved a 2 cm improvement in localization accuracy, a significant enhancement in this domain.
\end{abstract}



 % \includegraphics[width=\textwidth]{sampleteaser}
  %\caption{Seattle Mariners at Spring Training, 2010.}
  %\Description{Enjoying the baseball game from the third-base
  %seats. Ichiro Suzuki preparing to bat.}
  %\label{fig:teaser}

\maketitle
\section{INTRODUCTION}
Precise localization is a fundamental aspect of autonomous driving, supporting critical functions such as perception, path planning, and navigation, all of which are essential for ensuring safe and efficient operation.
\begin{figure}[h!]
\centering
  \includegraphics[width=\linewidth]{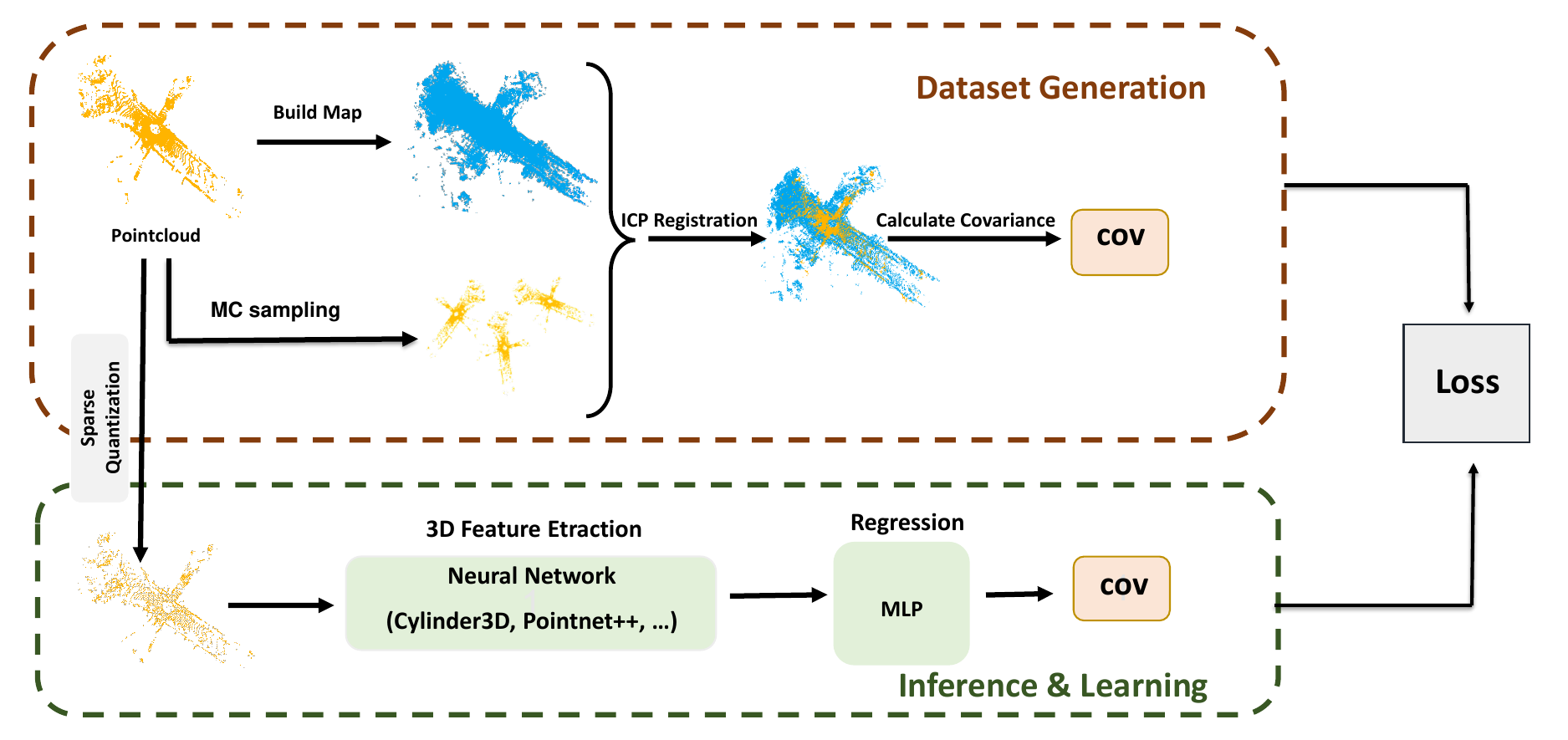}
  \caption[Proposed Approach Overview]{Overview of the proposed approach. In the upper section, a map is built from the input point cloud, followed by ICP matching to generate covariance for each point cloud. In the lower section, the input point cloud from the LiDAR sensor is processed through a neural network. The entire network, including feature extraction and regression, is trained end-to-end.}
  \label{our_net}
\end{figure}
In recent years, the application of Light Detection and Ranging (LiDAR) technology in autonomous vehicles (AVs) has gained widespread attention due to advancements in compactness, resolution, and cost efficiency. LiDAR-based localization techniques can be broadly classified into two main categories: SLAM and pre-built map-based approaches \cite{hao2023global}. While SLAM methods are effective for short distances, they are prone to accumulating errors over time \cite{chen2022overview}. In contrast, pre-built map-based approaches have demonstrated greater reliability. These approaches involve the offline generation of a point cloud map, which is then used during localization to match incoming sensor data, thereby determining the vehicle’s position within the map \cite{javanmardi2018factors}. Algorithms such as Iterative Closest Point (ICP) are frequently employed to perform map matching \cite{charroud2024localization}.

Despite its advantages, map-based localization faces significant challenges, such as environmental similarities and noise, which can introduce localization errors. Localization errors in autonomous vehicles arise from various factors, including the performance of matching algorithms, dynamic phenomena, input scans, and the accuracy of prebuilt maps. Sensor noise and uncertainties introduced by dynamic objects, combined with limitations in matching algorithms like ICP, can significantly reduce localization precision. Additionally, dynamic environmental changes and inaccuracies in prebuilt maps further exacerbate these challenges, necessitating robust techniques to manage these uncertainties and improve the reliability of localization processes \cite{javanmardi2020pre, brossard2020new, kan2021performance}. Addressing these challenges requires advanced methods such as sensor fusion, which rely on accurate error covariance estimation for each sensor \cite{brossard2020new}. While sensors like GPS and IMU benefit from well-established covariance models \cite{heroux2008principles}, techniques like ICP struggle to obtain accurate covariance estimates.

To address this limitation, we propose a data-driven approach utilizing deep learning to predict localization error covariance in prebuilt map frameworks. By leveraging advancements in deep learning techniques, our method delivers more accurate covariance predictions, thereby enhancing the precision of localization in autonomous vehicles. The pipeline of our proposed method is illustrated in Figure \ref{our_net}.

The structure of this paper is as follows: Section 2 reviews related work and existing methods for localization. Section 3 details our proposed method. Section 4 presents the experimental setup and results. Section 5 concludes with findings and future research directions.
\section{RELATED WORKS}
In the previous section, we stated our goal of estimating localization error covariance using a deep learning approach. Since no previous work has specifically addressed this topic, we analyze two relevant areas here. The first focuses on research related to predicting localization error, and the second examines studies that perform experiments involving covariance prediction.
\subsection{Predicting Localization Error}
Tuna et al. \cite{tuna2023x} highlight the ICP algorithm as the most widely used alignment method in LiDAR-based localization systems. However, they emphasize that the iterative minimization process can experience degeneration in complex environments, particularly those exhibiting geometric self-symmetry or lacking distinctive features. This degeneration often manifests as LiDAR drift when navigating similar pathways, such as narrow and long building corridors or tunnels, significantly impairing the accuracy of robot position estimation. To address these issues, various methods have been proposed to detect degenerative conditions by integrating them into uncertainty prediction models or position covariance estimation processes.

These methods can be broadly classified into three main approaches \cite{tuna2023x}: (1) \textit{geometric methods}, (2) \textit{optimization-based methods}, and (3) \textit{data-driven methods}.

\textbf{1. Geometric Methods:}
These methods focus on identifying degeneracies in maps and improving localization by analyzing the relationship between matching cost functions and the environment. Early work by Gelfand et al. \cite{gelfand2003geometrically} highlighted the instability of point-to-plane ICP in regions with indistinct features, proposing a sampling-based method for optimization. Kwok et al. \cite{kwok2016improvements} improved robustness with recalculated centers of mass and rotation normalization. Deschaud \cite{deschaud2018imls} introduced IMLS-SLAM, reducing drift by aligning scans with pre-built models, though it was computationally intensive. Zhen et al. \cite{zhen2019estimating} and Tuna et al. \cite{tuna2023x} further refined sensitivity analysis and localization by addressing LiDAR degeneration and scale differences, respectively, eliminating the need for pre-built maps and improving performance.

\textbf{2. Optimization-Based Methods:}  
Degeneracy detection has also been applied in various optimization and localization contexts, including SLAM. Rong et al. \cite{rong2016detection} utilized Gramian eigenvalues to assess the sufficiency of sensor measurements, while Ebadi et al. \cite{ebadi2021dare} and Cho et al. \cite{cho2018detection} employed the Hessian condition number to evaluate degeneracy in 6-DoF pose estimation. Liu et al. \cite{liu2021localizability} and Dong et al. \cite{dong2021efficient} suggested using the determinant of the Fisher Information Matrix for degeneracy detection, with Dong also introducing a PCA-based method to address feature deficiencies. Zhang et al. \cite{zhang2016degeneracy} introduced the concept of a degeneracy coefficient and proposed re-mapping strategies to mitigate optimization issues, which Hinduja et al. \cite{hinduja2019degeneracy} later refined using the Relative Condition Number for improved robustness.

\textbf{3. Data-Driven Methods:}  
Data-driven approaches have made significant strides in improving degeneracy detection and localization accuracy in ICP-based systems. Nobili et al. \cite{nobili2018predicting} introduced a method that integrates overlap data with a matchability metric, using support vector classification to predict alignment risks in environments with geometric constraints, such as flat surfaces. Adolfsson et al. \cite{adolfsson2021coral} extended this work by using entropy differences between point clouds to assess post-alignment accuracy and identify misalignments. Chen et al. \cite{chen2022overlapnet} developed OverlapNet, a Siamese neural network designed to estimate the overlap between lidar scans, improving the detection of environmental similarities.
Vega et al. \cite{vega2013cello} proposed CELLO, which replaces traditional covariance estimation by mapping point clouds into a descriptor space and using machine learning to predict uncertainty. This method was further extended by Landry et al. \cite{landry2019cello} with CELLO-3D, providing more accurate online covariance estimation for ICP systems in 3D environments. Nubert et al. \cite{nubert2022learning} contributed to this field by developing a method to estimate matching accuracy in unstructured environments through the comparison of expected and actual entropy values in shared point clouds, using a sparse 3D CNN based on ResUNet architecture.

In addition to these ICP advancements, notable work has been done in Normal Distribution Transform (NDT)-based localization.  Javanmardi et al. \cite{javanmardi2020pre} analyzed factors like feature sufficiency and layout to predict localization errors, showing that map accuracy alone is not enough for precise localization. They successfully predicted errors in urban environments by examining these map characteristics. Endo et al. \cite{endo2021analysis} further investigated how dynamic elements, such as moving vehicles, affect localization accuracy, offering insights into improving NDT-based localization in real-world, dynamic conditions.

These contributions to both ICP and NDT-based systems underscore the potential of data-driven methods to enhance alignment and localization accuracy. 
\subsection{Covariance Prediction}

A review of the literature revealed a gap in the use of neural networks for predicting localization error covariance, motivating our focus on this area. While covariance estimation has been applied in fields like position estimation \cite{liu2018deep} and time-series analysis \cite{fang2021cnn}, its potential in localization remains underexplored.

Deep learning approaches, such as Fang et al. \cite{fang2021cnn}, use CNNs and ConvLSTMs to predict covariance matrices in financial markets, providing flexibility and performance with large datasets. Liu et al. \cite{liu2018deep} applied CNNs to predict sensor measurement covariance for unmanned vehicles, offering real-time adaptability without relying on parametric noise models.

To ensure semi-positive definiteness of the covariance matrices, decomposition methods like Cholesky \cite{fang2021cnn, liu2018deep} and LDL \cite{fang2021cnn} are typically used, with our study adopting Cholesky. This research aims to extend these deep learning-based methods to predict localization error covariance, leveraging their effectiveness in three-dimensional data scenarios.

\section{Proposed Approach}
\subsection{ Pose Representation}
To facilitate a better understanding of the topics and discussions that follow, it is essential to introduce and explain some fundamental mathematical concepts.

A rigid transformation $\boldsymbol{T}$ between two point clouds, within the set of three-dimensional rigid transformations, is defined as: 
\begin{equation}
SE(3):=\left\{\left.\mathbf{T}=\left[\begin{array}{cc}
\mathbf{R} & \mathbf{t} \\
\mathbf{0} & 1
\end{array}\right] \in \mathbb{R}^{4 \times 4} \right\rvert\, \mathbf{R} \in S O(3), \mathbf{t} \in \mathbb{R}^3\right\}
\end{equation} 
where $\mathbf{R}$ represents a rotation matrix and $\mathbf{t}$ is a translation vector. 
Using Lie algebra, the transformation matrix $\boldsymbol{T}$ can be expressed as a vector $\boldsymbol{\xi} \in se(3)$ via the logarithm map $\log(\boldsymbol{T})$ \cite{blanco2021tutorial}. Conversely, the exponential map $\exp(\boldsymbol{\xi})$ reconstructs the transformation from the vector \cite{blanco2021tutorial}. The vector $\boldsymbol{\xi}$ is partitioned into a translation component $\boldsymbol{u} \in \mathbb{R}^3$ and an axis-angle rotation component $\boldsymbol{\omega} \in \mathbb{R}^3$, such that: 
\begin{equation} 
\boldsymbol{\xi}=\left[\begin{array}{l}
\boldsymbol{u} \\
\boldsymbol{\omega}
\end{array}\right] \quad \text { and } \quad \boldsymbol{Y}=\left[\begin{array}{ll}
\boldsymbol{Y}_{\boldsymbol{u} \boldsymbol{u}} & \boldsymbol{Y}_{\boldsymbol{u} \boldsymbol{\omega}} \\
\boldsymbol{Y}_{\boldsymbol{\omega} \boldsymbol{u}} & \boldsymbol{Y}_{\boldsymbol{\omega} \boldsymbol{\omega}}
\end{array}\right] ,
\end{equation}
where $\boldsymbol{Y} \in \mathbb{R}^{6 \times 6}$ is the covariance matrix representing the uncertainty in the rigid transformation.

In general, the ICP algorithm utilizes prior data or an initial transformation, denoted by $\boldsymbol{T}$, to obtain an estimate $\widehat{\boldsymbol{T}}$ of the true transformation $\overline{\boldsymbol{T}}$. For example, given a reference point cloud $\boldsymbol{Q} \in \mathbb{R}^3$ with $m$ points, we can compute an estimated transformation $\widehat{\boldsymbol{T}}$ that minimizes the alignment error between $\boldsymbol{P}$ and $\boldsymbol{Q}$ using an initial guess $\boldsymbol{T}$. A common solution to this alignment problem is to use the ICP algorithm: \begin{equation} \widehat{\boldsymbol{T}} = \operatorname{ICP}(\boldsymbol{P}, \boldsymbol{Q}, \boldsymbol{T}), \end{equation} where $\operatorname{ICP}(\cdot)$ denotes the application of the ICP algorithm to point clouds $\boldsymbol{P}$ and $\boldsymbol{Q}$ with the initial transformation $\boldsymbol{T}$.

The initial transformation $\boldsymbol{T}$ can be considered a random sample drawn from a distribution of transformations $\mathcal{O}$, whose shape typically depends on initial guesses provided by other sensors and the vehicle's velocity. Similarly, the estimated transformation $\widehat{\boldsymbol{T}}$ arises from the distribution $\mathcal{I}$, which has a complex form depending on both point clouds, the distribution $\mathcal{O}$, the configuration of the ICP algorithm $\operatorname{ICP}(\cdot)$, and the environments. 
In this work, the covariance estimate $\boldsymbol{Y}$ from ICP corresponds to the assumption that $\mathcal{I}$ is normally distributed.
\subsection{Overview}
The main goal of this research is to predict the error covariance in map-based localization techniques. While localization using pre-built maps provides accurate positioning, it is prone to errors caused by environmental similarities and sensor noise, which can reduce the system's reliability. To address these limitations, we propose a novel deep learning-based approach for predicting the error covariance in localization. An overview of the proposed method is presented in Figure \ref{our_net}.

The proposed model consists of two main components that work together to predict localization error covariance.
The first component introduces a novel Monte Carlo-based dataset generation method. This step involves creating a suitable dataset for predicting localization error covariance from pre-built maps. The dataset is designed to output a $6\times6$ error covariance matrix, representing uncertainty across various localization dimensions. This matrix is the ground truth for training and evaluating the deep learning model.
The second component employs a deep neural network framework to predict the error covariance. This part focuses on the development and training of a deep learning model capable of effectively predicting localization error covariance. The implementation of these components is detailed in the following sections.
\subsection{Dataset Generation Method}\label{generate-dataset}
This section outlines the preprocessing and dataset generation steps for training the neural network. A novel Monte Carlo-based map dataset generation method, inspired by \cite{nubert2022learning}, is introduced. Given that the ICP algorithm is sensitive to the initial guess, the process must be repeated $n$ times for each point cloud. Assuming we are in the $i$-th iteration for the $k$-th point cloud, the steps are as follows:

\begin{enumerate}
    \item \textbf{Map Construction}: After receiving a reference point cloud, a map is constructed using $x$ preceding and $y$ subsequent point clouds. The point clouds are transformed from their local coordinate frames to a global frame using ground truth transformations. This process is repeated for all point clouds to gradually build the map. If ground truth transformations are unavailable, mapping methods like LOAM \cite{zhang2014loam} can be used to generate accurate maps.
    
    \item \textbf{Data Reduction}: The input point cloud and the constructed map are reduced using common down-sampling techniques. In this study, voxel down-sampling with a voxel size of 1 meter for the map and 0.1 meter for the reference point cloud is used to reduce computational complexity and memory usage.
    
    \item \textbf{ICP Matching}: The point-to-plane ICP algorithm is applied to align the input point cloud with the constructed map. The initial transformation is computed as follows:
    \[
    \boldsymbol{T} = \exp (\boldsymbol{\xi}) \overline{\boldsymbol{T}},
    \]
    where $\boldsymbol{\xi} \sim \mathcal{N}(\mathbf{0}, \mathcal{O})$. This transformation $\exp (\boldsymbol{\xi})$ is converted into the SE(3) group to compute $\\boldsymbol{T}$. After applying ICP, the final transformation estimate $\widehat{T}$ is obtained.
    
    \item \textbf{Error Calculation}: The error is computed as:
    \[
    \boldsymbol{\xi}_i = \log \left(\overline{\boldsymbol{T}}_k^{-1} \widehat{\boldsymbol{T}}_i\right),
    \]
\end{enumerate}

After $n$ iterations, the covariance matrix is calculated as follows:
\[
\boldsymbol{Y}_k = \frac{1}{n-1} \sum_i \boldsymbol{\xi}_i \boldsymbol{\xi}_i^{\top}.
\]
where$ {Y}_k\in\mathbb{R}^{6 \times 6}$ denotes the estimation of the covariance matrix.
The pseudocode for the dataset generation process is shown in Algorithm\ref{alg-dataset}.
\begin{algorithm}[!ht]
    \caption{Dataset-Generation}\label{alg-dataset}
    \DontPrintSemicolon
    \LinesNumbered
    \KwIn{Set of point clouds}
    \KwOut{Covariance matrices $\boldsymbol{Y}_k$ for all $k$}
    
    \For{each point cloud $k$}{
        \For{$i \gets 1$ \KwTo $n$}{
            Construct map using $x$ previous and $y$ subsequent point clouds\;
            Transform point clouds from local to global coordinates using ground-truth\;
            Reduce data volume of the input point cloud and the constructed map\;
            Initialize $\boldsymbol{\xi} \sim \mathcal{N}(\mathbf{0}, \mathcal{O})$\;
            Compute $\boldsymbol{T} = \exp(\boldsymbol{\xi}) \overline{\boldsymbol{T}}$\;
            Apply ICP to align input point cloud with map, obtain estimate $\widehat{\boldsymbol{T}}$\;
            Compute $\boldsymbol{\xi}_i = \log\left(\overline{\boldsymbol{T}}_k^{-1} \widehat{\boldsymbol{T}}_i\right)$\;
        }
        Compute covariance matrix $\boldsymbol{Y}_k = \dfrac{1}{n-1} \sum_i \boldsymbol{\xi}_i \boldsymbol{\xi}_i^{\top}$\;
    }
    \Return Covariance matrices $\boldsymbol{Y}_k$ for all $k$\;
\end{algorithm}

\subsection{Training}
As shown in Figure \ref{our_net}, the proposed approach consists of two main components. In the first stage, the covariance of the dataset is calculated to capture uncertainties in the localization data. In the second stage, focused on feature extraction, the input point clouds are down-sampled to make the data more suitable for neural network processing. These down-sampled point clouds are then fed into neural networks such as Cylinder3D \cite{zhou2020cylinder3d} or PointNet++ \cite{Qi2017PointNetDH}.

Once the features are extracted, they are passed through a multi-layer regression network, which uses these features to make the final predictions.

Covariance matrices are inherently symmetric and positive definite. However, the outputs of the proposed network, which represent predicted covariance matrices, may not necessarily retain these properties. To ensure that the predicted covariance matrices remain symmetric and positive definite, we employ a Cholesky decomposition-based approach. In this method, a positive definite matrix \(\Sigma_t\) is expressed as \(\Sigma_t = C_t C_t^\top\), where \(C_t\) is a lower triangular matrix with positive, real diagonal entries. The output of our model is the lower triangular matrix \(C_t\), which is then converted to $\mathcal{\widehat{\boldsymbol{Y}}}$ using Cholesky decomposition. This guarantees that the predicted covariance matrices are always symmetric and positive definite.

 During each training iteration, samples are randomly selected from the dataset with replacement. Since most of the data points have near-zero covariance, weighted sampling is used, where the probability of selecting a sample is proportional to the absolute value of the largest element in its covariance matrix.

Data augmentation is also employed to maximize the use of computationally expensive samples. This process applies a transformation \(\boldsymbol{T}\) to the point clouds. If a transformation is applied to the \(k\)-th point cloud, the corresponding covariances are transformed using the adjoint representation of \(\boldsymbol{T}\), calculated as \cite{landry2019cello}:

\[
\overline{\boldsymbol{Y}} = \mathbf{A}_{\boldsymbol{T}}\boldsymbol{Y}\mathbf{A}_{\boldsymbol{T}}^{\top}.
\]

 In this study, data augmentation was performed by randomly translating the reference point cloud along the $x$ and $y$ axes and rotating it around the $z$ axis. This ensures the augmented data remains consistent with real-world conditions and avoids generating unrealistic samples. Other rotations could create samples with unnatural angles that do not accurately reflect the geometric features of the environment.

\subsection{Loss Function}\label{loss}
The objective is formulated as a regression problem for error covariance prediction. The combined loss function is defined as:

\begin{equation}
\mathcal{L}(\widehat{\boldsymbol{Y}}, \overline{\boldsymbol{Y}}) = \alpha D_{KL}[\mathcal{N}\left(0, \widehat{\boldsymbol{Y}}\right) \parallel \mathcal{N}\left(0, \overline{\boldsymbol{Y}}\right)] + \beta \sum_{i=1}^{n} L_{\text{Huber}}(\widehat{\boldsymbol{Y}}_{\text{up}, i} - \overline{\boldsymbol{Y}}_{\text{up}, i}),
\end{equation}
where \(\alpha = 0.1\) and \(\beta = 0.9\), which are empirically determined hyperparameters. Here, \(\widehat{\boldsymbol{Y}}\) represents the predicted covariance, and \(\overline{\boldsymbol{Y}}\) represents the ground truth covariance.\\
The Kullback-Leibler divergence is given by:
\begin{equation}
  D_{KL}[\mathcal{N}\left(0, \widehat{\boldsymbol{Y}}\right) \parallel \mathcal{N}\left(0, \overline{\boldsymbol{Y}}\right)] = 
  \frac{1}{2}\left(
\operatorname{tr}\left(\overline{\boldsymbol{Y}}^{-1}\widehat{\boldsymbol{Y}}\right)  - d +
    \ln\left(\frac{\det\overline{\boldsymbol{Y}}}{\det\widehat{\boldsymbol{Y}}}\right)
  \right).
  \label{d_kl}
\end{equation}

Where $d$ represent the dimension of the covariance matrix which is 6 in this case. The Huber loss is defined as:

\begin{equation}
L_{{\text{Huber}}}(\delta) =
\begin{cases}
\frac{1}{2} \delta^2 &  |\delta| \leq \delta_0,\\
\delta_0 (|\delta| - \frac{1}{2} \delta_0) &  |\delta| > \delta_0.
\end{cases}
\end{equation}

These loss functions have been commonly utilized in previous studies to enhance model performance. Since the covariance matrix is symmetric, the Huber loss is applied only to the upper (or lower) triangular elements.
\section{EXPERIMENT AND EVALUATION}
\subsection{Dataset Generation}
The proposed method generates new datasets using any LiDAR-based dataset. For this study, we utilized the KITTI Odometry dataset \cite{geiger2012we}, a subset of the  KITTI dataset, widely recognized for autonomous vehicle research. The dataset, captured in Germany with a Velodyne HDL-64E sensor, contains 22 sequences, with reference data available for the first 11 sequences.

The dataset generation process is as follows:
It begins with dataset selection, where a reference point cloud from the first 11 sequences is chosen to build a map. To ensure reliable results, the training, testing, and validation datasets were randomly selected, with 10,000, 5,000, and 2,500 samples, respectively. Following this, a map is constructed using the selected point cloud, which incorporates 20 previous and 10 subsequent point clouds relative to the input in the given sequence.

Once the map is constructed, a $6 \times 6$ covariance matrix is computed for each reference point cloud using the proposed method. The input point cloud is aligned with the constructed map through the ICP algorithm. This algorithm requires an initial transformation, sampled from a Gaussian distribution with a mean of zero and a specified covariance. The parameters for constructing the main diagonal of the initial covariance matrix are provided in Table \ref{parameters-our_work}, where $\sigma_x$, $\sigma_y$, $\sigma_z$ represent spatial uncertainty, and $\sigma_\phi$, $\sigma_\theta$, $\sigma_\psi$ represent rotational uncertainty.
    \begin{table}[h!] 
    \centering 
    \caption{Parameters for constructing the main diagonal of the initial covariance matrix.} 
    \begin{tabular}{|c|c|c|c|c|c|} 
    \hline 
     $\sigma_x[\mathrm{~m}]$ & $\sigma_y[\mathrm{~m}]$ & $\sigma_z[\mathrm{~m}]$ & $\sigma_\phi\left[^{\circ}\right]$ & $\sigma_\theta\left[^{\circ}\right]$ & $\sigma_\psi\left[^{\circ}\right]$ \\ \hline  
    1 & 1 & 1 & 5 & 5 & 5 \\ \hline 
    \end{tabular} 
    \label{parameters-our_work}
    \end{table}

\begin{figure}[!h]
      \centering
      \includegraphics[width=\linewidth]{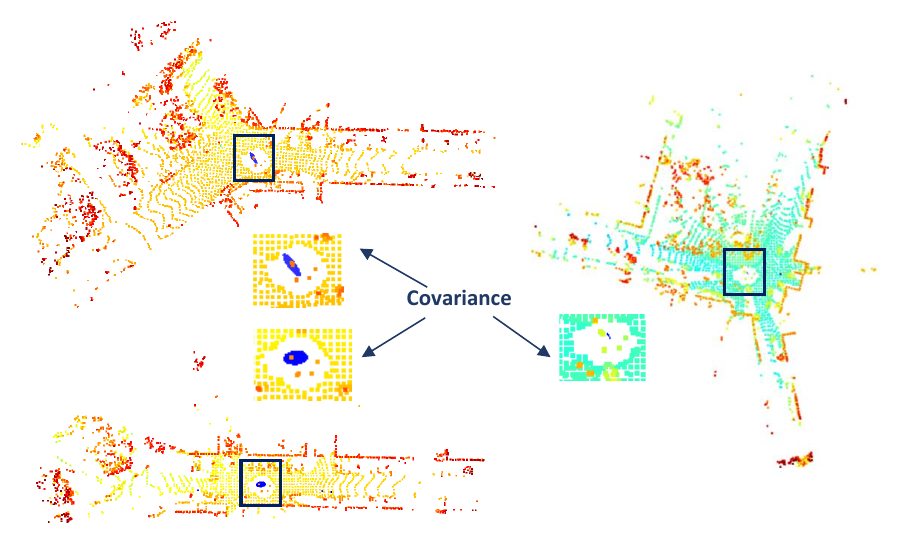}
      \caption{Visualization of computed covariance on the corresponding point cloud in sequence 00.}
      \label{covariance-on-point_cloud}
    \end{figure}

After 1000 iterations of alignment, the covariance for each input point cloud is calculated. Figure \ref{covariance-on-point_cloud} provides examples of the computed covariances, showing visualizations of point clouds at different stages of the sequence.
    
\subsection{Evaluation Metrics}
To evaluate the proposed approach, two metrics were applied, focusing on the upper (or lower) triangular elements of the covariance matrix due to its symmetry:

\begin{itemize}
    \item \textbf{Mean Absolute Error (MAE)}: Chosen for its simplicity and effectiveness.

    \item \textbf{Kullback-Leibler Divergence (KL)}: KL divergence measures the similarity between two distributions and is computed as shown in Equation \ref{d_kl}, averaged over the number of data in the evaluation datasets.
\end{itemize}

\subsection{Results}
In this section, we conducted experiments using different architectures based on the initial covariance values outlined in Table \ref{parameters-our_work}, and the results were thoroughly analyzed. First, we evaluated the proposed method on two networks, Cylinder3D \cite{zhou2020cylinder3d} and PointNet++ \cite{Qi2017PointNetDH}, using the KL divergence metric (Table \ref{table:kl-values}). Both networks demonstrated strong performance, though Cylinder3D outperformed PointNet++ due to its architecture, which efficiently combines 2D and 3D convolutions. This resulted in reduced computational complexity and enhanced accuracy in 3D data processing.

Additionally, we explored the use of pretrained weights from a semantic segmentation task on the SemanticKITTI dataset \cite{behley2019iccv}. As shown in Table \ref{table:kl-values}, this approach yielded the best performance. Alongside the KL divergence, the table also provides MAE evaluations for the x, y, and yaw.
\begin{table}[ht!]
%\centering
\caption{Results for KL and MAE metrics along the x, y axes, and yaw around the z axis for three neural networks.}
\resizebox{\columnwidth}{!}{%
\begin{tabular}{|c|c|c|c|c|c|}
%\midrule
\hline
\textbf{Backbone} & \textbf{Pretrained} & \textbf{KL} & \textbf{x (MAE)}& \textbf{y (MAE)}&\textbf{yaw (MAE)}\\ \hline
\textbf{Pointnet++} & $-$ & 7.435 & 0.1 & 0.05 & $2.8 \times 10^{-5}$ \\ \hline
\textbf{Cylinder3D} & $-$ & 6.612 & 0.04 & 0.07 & $2.6 \times 10^{-5}$ \\ \hline
\textbf{Cylinder3D} & \checkmark  & \textbf{6.173} & \textbf{0.03} & \textbf{0.02} &$\mathbf{9 \times 10^{-6}}$ \\ 
\hline
 %\bottomrule
\end{tabular}}
\label{table:kl-values}
\end{table}
\subsection{Evaluation of Sensor Fusion}
To assess the effectiveness of the proposed method, we integrated the predicted error covariance into a Kalman filter for real-time localization. The Kalman filter, a widely used algorithm for state estimation in dynamic systems, excels at handling noisy or uncertain sensor data by alternating between state prediction and correction using sensor measurements. In this study, we relied on Inertial Measurement Unit (IMU) data as the motion model, followed by corrections using ICP to achieve precise, real-time localization \cite{urrea2021kalman}. By incorporating the predicted covariance from our model into the Kalman filter at each stage, we evaluated the impact on localization accuracy and system stability. The results demonstrated notable improvements in both accuracy and robustness under real-world conditions.

For quantitative evaluation, we used one sequence from the KITTI-Odometry dataset. First, we computed the error covariances for this sequence and applied them within the Kalman filter framework. Three approaches were compared: one using fixed covariance (calculated as the average of the sequence’s covariances), another using dynamically updated covariances predicted by our model at each iteration, and a final approach using only ICP.

To evaluate performance, we used two standard metrics for trajectory prediction: Final Displacement Error (FDE) and Average Displacement Error (ADE). The results are shown in Table \ref{tab}.
\begin{table}[ht] 
\centering 
\caption{Comparison of ADE and FDE metrics.} 
\label{tab}
\begin{tabular}{|c|c|c|} 
\hline 
\textbf{Method} & \textbf{ADE (m)} & \textbf{FDE (m)} \\ \hline 
\text{Only ICP} & $0.05609$ & $0.05321$ \\ \hline
\text{Fixed Covariance + Kalman Filter} & $0.02998$ & $0.04127$ \\ \hline 
\text{Proposed Covariance + Kalman Filter} & $\mathbf{0.02097}$ & $\mathbf{0.02072}$\\ \hline 
\end{tabular} 
\end{table}

As shown, the Kalman filter utilizing the predicted covariance from the proposed method demonstrated significantly improved results. The fixed covariance used for comparison was the average of all predicted covariances for that sequence.
\section{CONCLUSION}
In this paper, we addressed the challenge of predicting localization error covariance in LiDAR-based map matching, a critical issue for improving the precision of autonomous vehicle localization systems. We proposed a novel deep learning-based framework that leverages a Monte Carlo-based dataset generation method to predict error covariance in prebuilt map frameworks.

Our results, based on the KITTI Odometry dataset, show that the proposed approach effectively improves localization accuracy. When integrated with a Kalman filter, the dynamically predicted covariance from our neural network enhanced the overall localization performance. The use of deep learning for predicting error covariance demonstrates its potential as a promising method for increasing the robustness of localization in autonomous vehicles.

 Future research could explore extending the framework to multi-sensor fusion scenarios or investigating the impact of different neural network architectures on performance in dynamic, real-world environments.
 

\bibliographystyle{ACM-Reference-Format}
\bibliography{sample-sigconf}


\begin{thebibliography}{36}

%%% the Web address.

\ifx \showCODEN    \undefined \def \showCODEN     #1{\unskip}     \fi
\ifx \showDOI      \undefined \def \showDOI       #1{#1}\fi
\ifx \showISBNx    \undefined \def \showISBNx     #1{\unskip}     \fi
\ifx \showISBNxiii \undefined \def \showISBNxiii  #1{\unskip}     \fi
\ifx \showISSN     \undefined \def \showISSN      #1{\unskip}     \fi
\ifx \showLCCN     \undefined \def \showLCCN      #1{\unskip}     \fi
\ifx \shownote     \undefined \def \shownote      #1{#1}          \fi
\ifx \showarticletitle \undefined \def \showarticletitle #1{#1}   \fi
\ifx \showURL      \undefined \def \showURL       {\relax}        \fi
% invisible to TeX
\providecommand\bibfield[2]{#2}
\providecommand\bibinfo[2]{#2}
\providecommand\natexlab[1]{#1}
\providecommand\showeprint[2][]{arXiv:#2}

\bibitem[Adolfsson et~al\mbox{.}(2021)]%
        {adolfsson2021coral}
\bibfield{author}{\bibinfo{person}{Daniel Adolfsson}, \bibinfo{person}{Martin Magnusson}, \bibinfo{person}{Qianfang Liao}, \bibinfo{person}{Achim~J Lilienthal}, {and} \bibinfo{person}{Henrik Andreasson}.} \bibinfo{year}{2021}\natexlab{}.
\newblock \showarticletitle{Coral--are the point clouds correctly aligned?}. In \bibinfo{booktitle}{\emph{2021 European conference on mobile robots (ECMR)}}. IEEE, \bibinfo{pages}{1--7}.
\newblock


\bibitem[Behley et~al\mbox{.}(2019)]%
        {behley2019iccv}
\bibfield{author}{\bibinfo{person}{J. Behley}, \bibinfo{person}{M. Garbade}, \bibinfo{person}{A. Milioto}, \bibinfo{person}{J. Quenzel}, \bibinfo{person}{S. Behnke}, \bibinfo{person}{C. Stachniss}, {and} \bibinfo{person}{J. Gall}.} \bibinfo{year}{2019}\natexlab{}.
\newblock \showarticletitle{{SemanticKITTI: A Dataset for Semantic Scene Understanding of LiDAR Sequences}}. In \bibinfo{booktitle}{\emph{Proc. of the IEEE/CVF International Conf.~on Computer Vision (ICCV)}}.
\newblock


\bibitem[Blanco-Claraco(2021)]%
        {blanco2021tutorial}
\bibfield{author}{\bibinfo{person}{Jos{\'e}~Luis Blanco-Claraco}.} \bibinfo{year}{2021}\natexlab{}.
\newblock \showarticletitle{A tutorial on SE(3) transformation parameterizations and on-manifold optimization}.
\newblock \bibinfo{journal}{\emph{arXiv preprint arXiv:2103.15980}} (\bibinfo{year}{2021}).
\newblock


\bibitem[Brossard et~al\mbox{.}(2020)]%
        {brossard2020new}
\bibfield{author}{\bibinfo{person}{Martin Brossard}, \bibinfo{person}{Silvere Bonnabel}, {and} \bibinfo{person}{Axel Barrau}.} \bibinfo{year}{2020}\natexlab{}.
\newblock \showarticletitle{A new approach to 3D ICP covariance estimation}.
\newblock \bibinfo{journal}{\emph{IEEE Robotics and Automation Letters}} \bibinfo{volume}{5}, \bibinfo{number}{2} (\bibinfo{year}{2020}), \bibinfo{pages}{744--751}.
\newblock


\bibitem[Charroud et~al\mbox{.}(2024)]%
        {charroud2024localization}
\bibfield{author}{\bibinfo{person}{Anas Charroud}, \bibinfo{person}{Karim El~Moutaouakil}, \bibinfo{person}{Vasile Palade}, \bibinfo{person}{Ali Yahyaouy}, \bibinfo{person}{Uche Onyekpe}, {and} \bibinfo{person}{Eyo~U Eyo}.} \bibinfo{year}{2024}\natexlab{}.
\newblock \showarticletitle{Localization and Mapping for Self-Driving Vehicles: A Survey}.
\newblock \bibinfo{journal}{\emph{Machines}} \bibinfo{volume}{12}, \bibinfo{number}{2} (\bibinfo{year}{2024}), \bibinfo{pages}{118}.
\newblock


\bibitem[Chen et~al\mbox{.}(2022b)]%
        {chen2022overview}
\bibfield{author}{\bibinfo{person}{Weifeng Chen}, \bibinfo{person}{Guangtao Shang}, \bibinfo{person}{Aihong Ji}, \bibinfo{person}{Chengjun Zhou}, \bibinfo{person}{Xiyang Wang}, \bibinfo{person}{Chonghui Xu}, \bibinfo{person}{Zhenxiong Li}, {and} \bibinfo{person}{Kai Hu}.} \bibinfo{year}{2022}\natexlab{b}.
\newblock \showarticletitle{An overview on visual slam: From tradition to semantic}.
\newblock \bibinfo{journal}{\emph{Remote Sensing}} \bibinfo{volume}{14}, \bibinfo{number}{13} (\bibinfo{year}{2022}), \bibinfo{pages}{3010}.
\newblock


\bibitem[Chen et~al\mbox{.}(2022a)]%
        {chen2022overlapnet}
\bibfield{author}{\bibinfo{person}{Xieyuanli Chen}, \bibinfo{person}{Thomas L{\"a}be}, \bibinfo{person}{Andres Milioto}, \bibinfo{person}{Timo R{\"o}hling}, \bibinfo{person}{Jens Behley}, {and} \bibinfo{person}{Cyrill Stachniss}.} \bibinfo{year}{2022}\natexlab{a}.
\newblock \showarticletitle{OverlapNet: A siamese network for computing LiDAR scan similarity with applications to loop closing and localization}.
\newblock \bibinfo{journal}{\emph{Autonomous Robots}} (\bibinfo{year}{2022}), \bibinfo{pages}{1--21}.
\newblock


\bibitem[Cho et~al\mbox{.}(2018)]%
        {cho2018detection}
\bibfield{author}{\bibinfo{person}{HyunGi Cho}, \bibinfo{person}{Suyong Yeon}, \bibinfo{person}{Hyunga Choi}, {and} \bibinfo{person}{Nakju Doh}.} \bibinfo{year}{2018}\natexlab{}.
\newblock \showarticletitle{Detection and compensation of degeneracy cases for imu-kinect integrated continuous slam with plane features}.
\newblock \bibinfo{journal}{\emph{Sensors}} \bibinfo{volume}{18}, \bibinfo{number}{4} (\bibinfo{year}{2018}), \bibinfo{pages}{935}.
\newblock


\bibitem[Deschaud(2018)]%
        {deschaud2018imls}
\bibfield{author}{\bibinfo{person}{Jean-Emmanuel Deschaud}.} \bibinfo{year}{2018}\natexlab{}.
\newblock \showarticletitle{IMLS-SLAM: Scan-to-model matching based on 3D data}. In \bibinfo{booktitle}{\emph{2018 IEEE International Conference on Robotics and Automation (ICRA)}}. IEEE, \bibinfo{pages}{2480--2485}.
\newblock


\bibitem[Dong et~al\mbox{.}(2021)]%
        {dong2021efficient}
\bibfield{author}{\bibinfo{person}{Lingfeng Dong}, \bibinfo{person}{Weidong Chen}, {and} \bibinfo{person}{Jingchuan Wang}.} \bibinfo{year}{2021}\natexlab{}.
\newblock \showarticletitle{Efficient feature extraction and localizability based matching for lidar slam}. In \bibinfo{booktitle}{\emph{2021 IEEE International Conference on Robotics and Biomimetics (ROBIO)}}. IEEE, \bibinfo{pages}{820--825}.
\newblock


\bibitem[Ebadi et~al\mbox{.}(2021)]%
        {ebadi2021dare}
\bibfield{author}{\bibinfo{person}{Kamak Ebadi}, \bibinfo{person}{Matteo Palieri}, \bibinfo{person}{Sally Wood}, \bibinfo{person}{Curtis Padgett}, {and} \bibinfo{person}{Ali-akbar Agha-mohammadi}.} \bibinfo{year}{2021}\natexlab{}.
\newblock \showarticletitle{DARE-SLAM: Degeneracy-aware and resilient loop closing in perceptually-degraded environments}.
\newblock \bibinfo{journal}{\emph{Journal of Intelligent \& Robotic Systems}}  \bibinfo{volume}{102} (\bibinfo{year}{2021}), \bibinfo{pages}{1--25}.
\newblock


\bibitem[Endo et~al\mbox{.}(2021)]%
        {endo2021analysis}
\bibfield{author}{\bibinfo{person}{Yuki Endo}, \bibinfo{person}{Ehsan Javanmardi}, {and} \bibinfo{person}{Shunsuke Kamijo}.} \bibinfo{year}{2021}\natexlab{}.
\newblock \showarticletitle{Analysis of occlusion effects for map-based self-localization in urban areas}.
\newblock \bibinfo{journal}{\emph{Sensors}} \bibinfo{volume}{21}, \bibinfo{number}{15} (\bibinfo{year}{2021}), \bibinfo{pages}{5196}.
\newblock


\bibitem[Fang et~al\mbox{.}(2021)]%
        {fang2021cnn}
\bibfield{author}{\bibinfo{person}{Yanwen Fang}, \bibinfo{person}{Philip~LH Yu}, {and} \bibinfo{person}{Yaohua Tang}.} \bibinfo{year}{2021}\natexlab{}.
\newblock \showarticletitle{CNN-based realized covariance matrix forecasting}.
\newblock \bibinfo{journal}{\emph{arXiv preprint arXiv:2107.10602}} (\bibinfo{year}{2021}).
\newblock


\bibitem[Geiger et~al\mbox{.}(2012)]%
        {geiger2012we}
\bibfield{author}{\bibinfo{person}{Andreas Geiger}, \bibinfo{person}{Philip Lenz}, {and} \bibinfo{person}{Raquel Urtasun}.} \bibinfo{year}{2012}\natexlab{}.
\newblock \showarticletitle{Are we ready for autonomous driving? the kitti vision benchmark suite}. In \bibinfo{booktitle}{\emph{2012 IEEE conference on computer vision and pattern recognition}}. IEEE, \bibinfo{pages}{3354--3361}.
\newblock


\bibitem[Gelfand et~al\mbox{.}(2003)]%
        {gelfand2003geometrically}
\bibfield{author}{\bibinfo{person}{Natasha Gelfand}, \bibinfo{person}{Leslie Ikemoto}, \bibinfo{person}{Szymon Rusinkiewicz}, {and} \bibinfo{person}{Marc Levoy}.} \bibinfo{year}{2003}\natexlab{}.
\newblock \showarticletitle{Geometrically stable sampling for the ICP algorithm}. In \bibinfo{booktitle}{\emph{Fourth International Conference on 3-D Digital Imaging and Modeling, 2003. 3DIM 2003. Proceedings.}} IEEE, \bibinfo{pages}{260--267}.
\newblock


\bibitem[Hao et~al\mbox{.}(2023)]%
        {hao2023global}
\bibfield{author}{\bibinfo{person}{Yun Hao}, \bibinfo{person}{Jiacheng Liu}, \bibinfo{person}{Yuzhen Liu}, \bibinfo{person}{Xinyuan Liu}, \bibinfo{person}{Ziyang Meng}, {and} \bibinfo{person}{Fei Xing}.} \bibinfo{year}{2023}\natexlab{}.
\newblock \showarticletitle{Global Visual--Inertial Localization for Autonomous Vehicles with Pre-Built Map}.
\newblock \bibinfo{journal}{\emph{Sensors}} \bibinfo{volume}{23}, \bibinfo{number}{9} (\bibinfo{year}{2023}), \bibinfo{pages}{4510}.
\newblock


\bibitem[Heroux(2008)]%
        {heroux2008principles}
\bibfield{author}{\bibinfo{person}{Pierre Heroux}.} \bibinfo{year}{2008}\natexlab{}.
\newblock \showarticletitle{Principles of GNSS, Inertial, and Multisensor Integrated Navigation Systems.}
\newblock \bibinfo{journal}{\emph{Geomatica}} \bibinfo{volume}{62}, \bibinfo{number}{2} (\bibinfo{year}{2008}), \bibinfo{pages}{207--209}.
\newblock


\bibitem[Hinduja et~al\mbox{.}(2019)]%
        {hinduja2019degeneracy}
\bibfield{author}{\bibinfo{person}{Akshay Hinduja}, \bibinfo{person}{Bing-Jui Ho}, {and} \bibinfo{person}{Michael Kaess}.} \bibinfo{year}{2019}\natexlab{}.
\newblock \showarticletitle{Degeneracy-aware factors with applications to underwater slam}. In \bibinfo{booktitle}{\emph{2019 IEEE/RSJ International Conference on Intelligent Robots and Systems (IROS)}}. IEEE, \bibinfo{pages}{1293--1299}.
\newblock


\bibitem[Javanmardi et~al\mbox{.}(2018)]%
        {javanmardi2018factors}
\bibfield{author}{\bibinfo{person}{Ehsan Javanmardi}, \bibinfo{person}{Mahdi Javanmardi}, \bibinfo{person}{Yanlei Gu}, {and} \bibinfo{person}{Shunsuke Kamijo}.} \bibinfo{year}{2018}\natexlab{}.
\newblock \showarticletitle{Factors to evaluate capability of map for vehicle localization}.
\newblock \bibinfo{journal}{\emph{IEEE Access}}  \bibinfo{volume}{6} (\bibinfo{year}{2018}), \bibinfo{pages}{49850--49867}.
\newblock


\bibitem[Javanmardi et~al\mbox{.}(2020)]%
        {javanmardi2020pre}
\bibfield{author}{\bibinfo{person}{Ehsan Javanmardi}, \bibinfo{person}{Mahdi Javanmardi}, \bibinfo{person}{Yanlei Gu}, {and} \bibinfo{person}{Shunsuke Kamijo}.} \bibinfo{year}{2020}\natexlab{}.
\newblock \showarticletitle{Pre-estimating self-localization error of NDT-based map-matching from map only}.
\newblock \bibinfo{journal}{\emph{IEEE Transactions on Intelligent Transportation Systems}} \bibinfo{volume}{22}, \bibinfo{number}{12} (\bibinfo{year}{2020}), \bibinfo{pages}{7652--7666}.
\newblock


\bibitem[Kan et~al\mbox{.}(2021)]%
        {kan2021performance}
\bibfield{author}{\bibinfo{person}{Yin-Chiu Kan}, \bibinfo{person}{Li-Ta Hsu}, {and} \bibinfo{person}{Edward Chung}.} \bibinfo{year}{2021}\natexlab{}.
\newblock \showarticletitle{Performance evaluation on map-based NDT scan matching localization using simulated occlusion datasets}.
\newblock \bibinfo{journal}{\emph{IEEE Sensors Letters}} \bibinfo{volume}{5}, \bibinfo{number}{3} (\bibinfo{year}{2021}), \bibinfo{pages}{1--4}.
\newblock


\bibitem[Kwok and Tang(2016)]%
        {kwok2016improvements}
\bibfield{author}{\bibinfo{person}{Tsz-Ho Kwok} {and} \bibinfo{person}{Kai Tang}.} \bibinfo{year}{2016}\natexlab{}.
\newblock \showarticletitle{Improvements to the iterative closest point algorithm for shape registration in manufacturing}.
\newblock \bibinfo{journal}{\emph{Journal of Manufacturing Science and Engineering}} \bibinfo{volume}{138}, \bibinfo{number}{1} (\bibinfo{year}{2016}), \bibinfo{pages}{011014}.
\newblock


\bibitem[Landry et~al\mbox{.}(2019)]%
        {landry2019cello}
\bibfield{author}{\bibinfo{person}{David Landry}, \bibinfo{person}{Fran{\c{c}}ois Pomerleau}, {and} \bibinfo{person}{Philippe Giguere}.} \bibinfo{year}{2019}\natexlab{}.
\newblock \showarticletitle{CELLO-3D: Estimating the Covariance of ICP in the Real World}. In \bibinfo{booktitle}{\emph{2019 International Conference on Robotics and Automation (ICRA)}}. IEEE, \bibinfo{pages}{8190--8196}.
\newblock


\bibitem[Liu et~al\mbox{.}(2018)]%
        {liu2018deep}
\bibfield{author}{\bibinfo{person}{Katherine Liu}, \bibinfo{person}{Kyel Ok}, \bibinfo{person}{William Vega-Brown}, {and} \bibinfo{person}{Nicholas Roy}.} \bibinfo{year}{2018}\natexlab{}.
\newblock \showarticletitle{Deep inference for covariance estimation: Learning gaussian noise models for state estimation}. In \bibinfo{booktitle}{\emph{2018 IEEE International Conference on Robotics and Automation (ICRA)}}. IEEE, \bibinfo{pages}{1436--1443}.
\newblock


\bibitem[Liu et~al\mbox{.}(2021)]%
        {liu2021localizability}
\bibfield{author}{\bibinfo{person}{Ying Liu}, \bibinfo{person}{Jingchuan Wang}, {and} \bibinfo{person}{Yi Huang}.} \bibinfo{year}{2021}\natexlab{}.
\newblock \showarticletitle{A localizability estimation method for mobile robots based on 3d point cloud feature}. In \bibinfo{booktitle}{\emph{2021 IEEE International Conference on Real-time Computing and Robotics (RCAR)}}. IEEE, \bibinfo{pages}{1035--1041}.
\newblock


\bibitem[Nobili et~al\mbox{.}(2018)]%
        {nobili2018predicting}
\bibfield{author}{\bibinfo{person}{Simona Nobili}, \bibinfo{person}{Georgi Tinchev}, {and} \bibinfo{person}{Maurice Fallon}.} \bibinfo{year}{2018}\natexlab{}.
\newblock \showarticletitle{Predicting alignment risk to prevent localization failure}. In \bibinfo{booktitle}{\emph{2018 IEEE International Conference on Robotics and Automation (ICRA)}}. IEEE, \bibinfo{pages}{1003--1010}.
\newblock


\bibitem[Nubert et~al\mbox{.}(2022)]%
        {nubert2022learning}
\bibfield{author}{\bibinfo{person}{Julian Nubert}, \bibinfo{person}{Etienne Walther}, \bibinfo{person}{Shehryar Khattak}, {and} \bibinfo{person}{Marco Hutter}.} \bibinfo{year}{2022}\natexlab{}.
\newblock \showarticletitle{Learning-based localizability estimation for robust lidar localization}. In \bibinfo{booktitle}{\emph{2022 IEEE/RSJ International Conference on Intelligent Robots and Systems (IROS)}}. IEEE, \bibinfo{pages}{17--24}.
\newblock


\bibitem[Qi et~al\mbox{.}(2017)]%
        {Qi2017PointNetDH}
\bibfield{author}{\bibinfo{person}{C. Qi}, \bibinfo{person}{L. Yi}, \bibinfo{person}{Hao Su}, {and} \bibinfo{person}{Leonidas~J. Guibas}.} \bibinfo{year}{2017}\natexlab{}.
\newblock \showarticletitle{PointNet++: Deep Hierarchical Feature Learning on Point Sets in a Metric Space}. In \bibinfo{booktitle}{\emph{Neural Information Processing Systems}}.
\newblock
\urldef\tempurl%
\url{https://api.semanticscholar.org/CorpusID:1745976}
\showURL{%
\tempurl}


\bibitem[Rong and Michael(2016)]%
        {rong2016detection}
\bibfield{author}{\bibinfo{person}{Zheng Rong} {and} \bibinfo{person}{Nathan Michael}.} \bibinfo{year}{2016}\natexlab{}.
\newblock \showarticletitle{Detection and prediction of near-term state estimation degradation via online nonlinear observability analysis}. In \bibinfo{booktitle}{\emph{2016 IEEE International Symposium on Safety, Security, and Rescue Robotics (SSRR)}}. IEEE, \bibinfo{pages}{28--33}.
\newblock


\bibitem[Tuna et~al\mbox{.}(2023)]%
        {tuna2023x}
\bibfield{author}{\bibinfo{person}{Turcan Tuna}, \bibinfo{person}{Julian Nubert}, \bibinfo{person}{Yoshua Nava}, \bibinfo{person}{Shehryar Khattak}, {and} \bibinfo{person}{Marco Hutter}.} \bibinfo{year}{2023}\natexlab{}.
\newblock \showarticletitle{X-icp: Localizability-aware lidar registration for robust localization in extreme environments}.
\newblock \bibinfo{journal}{\emph{IEEE Transactions on Robotics}} (\bibinfo{year}{2023}).
\newblock


\bibitem[Urrea and Agramonte(2021)]%
        {urrea2021kalman}
\bibfield{author}{\bibinfo{person}{Claudio Urrea} {and} \bibinfo{person}{Rayko Agramonte}.} \bibinfo{year}{2021}\natexlab{}.
\newblock \showarticletitle{Kalman filter: historical overview and review of its use in robotics 60 years after its creation}.
\newblock \bibinfo{journal}{\emph{Journal of Sensors}} \bibinfo{volume}{2021}, \bibinfo{number}{1} (\bibinfo{year}{2021}), \bibinfo{pages}{9674015}.
\newblock


\bibitem[Vega-Brown et~al\mbox{.}(2013)]%
        {vega2013cello}
\bibfield{author}{\bibinfo{person}{William Vega-Brown}, \bibinfo{person}{Abraham Bachrach}, \bibinfo{person}{Adam Bry}, \bibinfo{person}{Jonathan Kelly}, {and} \bibinfo{person}{Nicholas Roy}.} \bibinfo{year}{2013}\natexlab{}.
\newblock \showarticletitle{Cello: A fast algorithm for covariance estimation}. In \bibinfo{booktitle}{\emph{2013 IEEE International Conference on Robotics and Automation}}. IEEE, \bibinfo{pages}{3160--3167}.
\newblock


\bibitem[Zhang et~al\mbox{.}(2016)]%
        {zhang2016degeneracy}
\bibfield{author}{\bibinfo{person}{Ji Zhang}, \bibinfo{person}{Michael Kaess}, {and} \bibinfo{person}{Sanjiv Singh}.} \bibinfo{year}{2016}\natexlab{}.
\newblock \showarticletitle{On degeneracy of optimization-based state estimation problems}. In \bibinfo{booktitle}{\emph{2016 IEEE international conference on robotics and automation (ICRA)}}. IEEE, \bibinfo{pages}{809--816}.
\newblock


\bibitem[Zhang et~al\mbox{.}(2014)]%
        {zhang2014loam}
\bibfield{author}{\bibinfo{person}{Ji Zhang}, \bibinfo{person}{Sanjiv Singh}, {et~al\mbox{.}}} \bibinfo{year}{2014}\natexlab{}.
\newblock \showarticletitle{LOAM: Lidar odometry and mapping in real-time.}. In \bibinfo{booktitle}{\emph{Robotics: Science and systems}}, Vol.~\bibinfo{volume}{2}. Berkeley, CA, \bibinfo{pages}{1--9}.
\newblock


\bibitem[Zhen and Scherer(2019)]%
        {zhen2019estimating}
\bibfield{author}{\bibinfo{person}{Weikun Zhen} {and} \bibinfo{person}{Sebastian Scherer}.} \bibinfo{year}{2019}\natexlab{}.
\newblock \showarticletitle{Estimating the localizability in tunnel-like environments using LiDAR and UWB}. In \bibinfo{booktitle}{\emph{2019 International Conference on Robotics and Automation (ICRA)}}. IEEE, \bibinfo{pages}{4903--4908}.
\newblock


\bibitem[Zhou et~al\mbox{.}(2020)]%
        {zhou2020cylinder3d}
\bibfield{author}{\bibinfo{person}{Hui Zhou}, \bibinfo{person}{Xinge Zhu}, \bibinfo{person}{Xiao Song}, \bibinfo{person}{Yuexin Ma}, \bibinfo{person}{Zhe Wang}, \bibinfo{person}{Hongsheng Li}, {and} \bibinfo{person}{Dahua Lin}.} \bibinfo{year}{2020}\natexlab{}.
\newblock \showarticletitle{Cylinder3d: An effective 3d framework for driving-scene lidar semantic segmentation}.
\newblock \bibinfo{journal}{\emph{arXiv preprint arXiv:2008.01550}} (\bibinfo{year}{2020}).
\newblock


\end{thebibliography}
\end{document}